\documentclass[10pt,twocolumn,letterpaper]{article}

\usepackage[pagenumbers]{cvpr} % To force page numbers, e.g. for an arXiv version

\usepackage{amssymb}
\usepackage{textcomp}
\usepackage{bbding}
\usepackage{multirow}

\usepackage{marvosym}

\definecolor{cvprblue}{rgb}{0.21,0.49,0.74}
\usepackage[pagebackref,breaklinks,colorlinks,allcolors=cvprblue]{hyperref}

\def\paperID{***} % *** Enter the Paper ID here
\def\confName{CVPR}
\def\confYear{2026}

\title{Pressure2Motion: Hierarchical Human Motion Reconstruction
\\from Ground Pressure with Text Guidance}

\author{
    Zhengxuan Li\textsuperscript{\rm 1} \quad
    Qinhui Yang\textsuperscript{\rm 1} \quad
    Yiyu Zhuang\textsuperscript{\rm 1} \quad
    Chuan Guo\textsuperscript{\rm 2} \quad
    Xinxin Zuo\textsuperscript{\rm 3} \\
    Xiaoxiao Long\textsuperscript{\rm 1} \quad
    Yao Yao\textsuperscript{\rm 1} \quad
    Xun Cao\textsuperscript{\rm 1} \quad
    Qiu Shen\textsuperscript{\rm 1} \quad
    Hao Zhu\textsuperscript{\rm 1,\Letter}
    \vspace{0.5em}
    \\
    \textsuperscript{\rm 1}Nanjing University \quad
    \textsuperscript{\rm 2}Snap Inc. \quad
    \textsuperscript{\rm 3}Concordia University
    \vspace{-0.5em}
}

\usepackage{color}
\definecolor{myPurple}{rgb}{0.4, .0, .8}
\definecolor{myGreen}{rgb}{0, .8, .3}
\definecolor{myRed}{rgb}{0.8, .2, .2}
\definecolor{myOrange}{rgb}{0.8, 0.45, 0.0}
\definecolor{myBlue}{rgb}{.0, .0, 1.0}

\definecolor{expyellow}{RGB}{192,138,29}
\definecolor{expblue}{RGB}{90,144,193}

\newcommand{\expyellow}[1]{\textcolor{expyellow}{#1}}
\newcommand{\expblue}[1]{\textcolor{expblue}{#1}}

\newcommand{\methodname}{{\textit{Pressure2Motion}}}
\newcommand{\modelname}{{\textit{Pressure2Motion} }}
\newcommand{\dataname}{{\textit{MPL} }}

\begin{document}

\twocolumn[{
\renewcommand\twocolumn[1][!htb]{#1}%
\maketitle
\begin{center}
    \centering
    \includegraphics[width=1.0\linewidth]{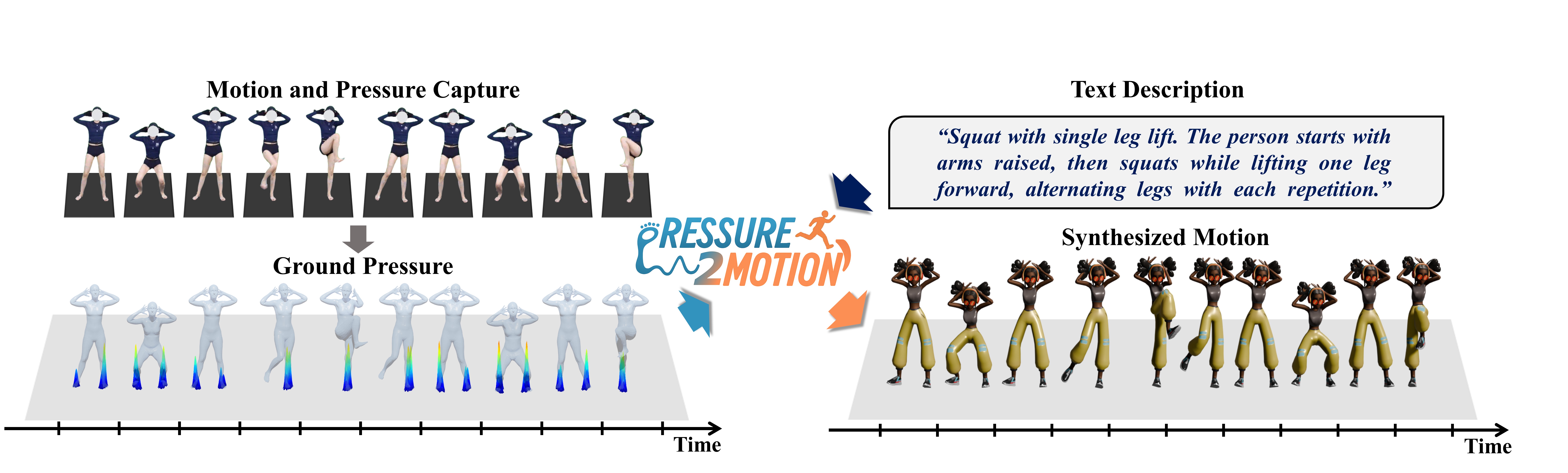}
    \captionof{figure}{By conditioning on pressure signals and text descriptions, Pressure2Motion reconstructs high-fidelity, physically realistic motions, addressing the challenge of synthesizing human motion from sparse and noisy pressure data.}
    \label{fig:teaser}
\end{center}
}]

\begin{abstract}

We present \methodname, a novel motion capture algorithm that reconstructs human motion from a ground pressure sequence and text prompt.  
At inference time, \modelname
requires only a pressure mat, eliminating the need for specialized lighting setups, cameras, or wearable devices, making it suitable for privacy-preserving, low-light, and low-cost motion capture scenarios.
Such a task is severely ill-posed due to the indeterminacy of pressure signals with respect to full-body motion.
To address this issue, we introduce \methodname, a generative model that leverages pressure features as input and utilizes a text prompt as a high-level guiding constraint to resolve ambiguities.
Specifically, our model adopts a dual-level feature extractor to accurately interpret pressure data, followed by a hierarchical diffusion model that discerns broad-scale movement trajectories and subtle posture adjustments.
Both the physical cues gained from the pressure sequence and the semantic guidance derived from descriptive texts are leveraged to guide the motion estimation with precision.
To the best of our knowledge, \modelname is a pioneering work in leveraging both pressure data and linguistic priors for motion reconstruction, and the established \dataname benchmark is the first benchmark for this novel motion capture task.
Experiments show that our method generates high-fidelity, physically plausible motions, establishing a new state of the art for this task.
The codes and benchmarks will be publicly released upon publication.

\end{abstract}
    
\section{Introduction}
\label{sec:intro}

Motion capture (MoCap) is a pivotal technology in digital animation production and robot control.  However, classical motion capture systems are plagued by numerous limitations, such as reliance on wearable devices (in optical MoCap~\cite{kirk2004skeletal, guerra2005optical} and inertial MoCap~\cite{cloete2008benchmarking, vitali2020determining}), the high cost of equipment (particularly for optical motion capture), and dependence on illumination and visual information (in visual MoCap~\cite{mehta2020xnect}).  These limitations have raised concerns regarding the cost, efficiency, and visual privacy protection of motion capture.  
%To address this issue, we propose employing ground pressure mats in conjunction with simple textual prompts for motion capture, thereby eliminating the reliance on illumination and wearable devices.

To address these limitations, we propose a novel MoCap paradigm: an algorithm that, at inference time, reconstructs full-body 3D motion using only a ground pressure mat in conjunction with simple textual prompts. Our goal is to create a system that can be deployed in privacy-sensitive environments without cameras or the need for users to wear invasive sensors. To learn the complex, ill-posed mapping from sparse pressure signals to full-body motion, our system is trained on a multimodal dataset containing synchronized pressure signals, ground-truth motion (captured via traditional sensors), and text. However, once trained, our system operates for inference without requiring any visual data or wearable devices.

However, reconstructing human motion from ground pressure data is a highly challenging task due to its inherently underdetermined nature. Previous work on human pose prediction from pressure data has achieved success only in scenarios with large contact areas, such as predicting the poses of individuals lying in bed \cite{clever2020bodies, wu2024seeing}. Regressing human motions in standing poses, such as walking, from ground pressure data presents a far greater challenge. 
On the other hand, although significant progress has been made in the text-to-motion task~\cite{guo2020action2motion, t2m, motiondiffuse, tevet2023human, petrovich2022temos, jiang2023motiongpt, chen2024text, momask}, using text prompts as control signals is too unconstrained for a MoCap system. 
More advanced controllable synthesis methods~\cite{posenetedit, karunratanakul2023guided, everything2motion, omnicontrol, Zhong2025Sketch2Anim, Pinyoanuntapong2025MaskControl, meng2025absolute, Hwang_2025_ICCV} offer finer control but rely on clean, manually-specified, and kinematically abstract inputs (e.g., trajectories or keypoints) that have a direct geometric correspondence to the desired motion. Their architectures are ill-equipped to handle a physically grounded, noisy, and indirect control modality such as ground pressure, which lacks a simple kinematic mapping to full-body pose.

In this paper, we propose \methodname, a novel generative model that, for the first time, conditions motion estimation % synthesis 
on both high-level semantic text and low-level physical pressure as shown in \cref{fig:teaser}. 
Our framework introduces a \textit{Dual-level Pressure Feature Extractor} that interprets both the rich semantics of pressure (e.g., identifying which body part is in contact, its direction, and magnitude) and its temporal dynamics. This extractor decodes pressure sequences into two distinct control signals: an overall \textit{Pressure-Inferred Movement Trajectory} and fine-grained \textit{Pressure-Induced Posture Shifts}. These signals are then injected into a pre-trained motion diffusion model via a novel \textit{Hierarchical Pressure-Modulated Motion Synthesizer}, which uses a ControlNet and a parallel Adapter module to ensure the final motion is both plausible and semantically consistent.

To advance this novel research direction, we establish the \dataname dataset. Constructed upon the extensive MotionPRO dataset~\cite{ren2025motionpro}. \dataname incorporates fresh, meticulously detailed textual annotations for each motion sequence, thereby establishing the inaugural large-scale, paired (text, pressure, motion) benchmark tailored specifically for this research task. We rigorously evaluate our proposed methodology on this dataset, and comprehensive experimental results demonstrate that \modelname can reconstruct human motions that are both highly realistic and physically coherent. This achievement heralds a new paradigm in non-visual, privacy-preserving motion capture.

Our contributions can be summarized as follows:
\begin{itemize}
    \item To the best of our knowledge, we are the first to formulate and address the novel task of % generating 
    full-body human motion capture from the highly sparse inputs of ground pressure and text prompts.
    \item A novel hierarchical generative network is designed to extract multi-level features from the ground pressure signals, which are injected hierarchically into a diffusion model for faithful motion synthesis.
    \item We demonstrate how to effectively adapt a pretrained text-to-motion model for this novel MoCap task, achieving both spatial alignment with ground pressure and semantic plausibility.
    \item We introduce the \dataname dataset, the first large-scale benchmark with paired text, pressure, and motion data, and demonstrate that our method achieves state-of-the-art performance on the benchmark.
\end{itemize}

\section{Related Work}
\label{sec:related}
\subsection{Pressure-based Pose Estimation}

\begin{figure*}[th]
\centering
\includegraphics[width=1.0\textwidth]{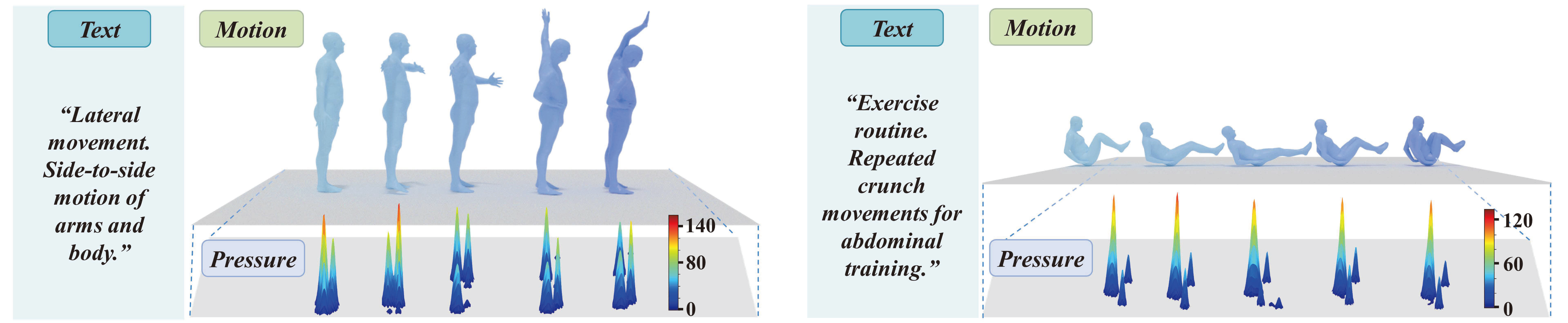}
\vspace{-0.2in}
\caption{Samples from our MPL dataset, each pairing ground pressure and motion with five levels of text descriptions.}
\vspace{-0.15in}
\label{fig:dataset}
\end{figure*}

Traditional pose estimation methods rely on RGB cameras or inertial measurement units (IMUs)\cite{martinez2017simple, pavllo20193d, von2017sparse, huang2018deep, joo2018total}, which lack reliable foot-ground contact cues and are unsuitable for privacy-sensitive scenarios like healthcare. 
Early pressure-based methods~\cite{yousefi2011bed,davoodnia2021bed,clever2020bodies} used single-frame pressure maps to estimate static postures, but were limited to predefined lying poses and low-resolution settings. Recent methods like PIMesh~\cite{wu2024seeing} extend this to short pressure sequences for mesh regression but struggle with dynamic motions like walking.
Derived signals like center of pressure (COP)\cite{funk2018learning,tripathi20233d} and foot contact\cite{zhang2024mmvp} offer stability cues but are limited to quasi-static scenarios. Existing pressure datasets are mainly focused on in-bed or insole data, limiting coverage for free-moving activities. Most methods are constrained to short, static actions, with long dynamic sequences remaining a challenge.

Multi-modal approaches, such as BodyPressure~\cite{clever2022bodypressure} and PressInPose~\cite{gao2024pressinpose}, and others~\cite{yin2022multimodal,liu2022simultaneously,tandon2024bodymap} combine pressure with RGB, depth, or thermal sensors to improve pose accuracy, but still rely heavily on cameras, compromising privacy. MotionPRO~\cite{ren2025motionpro} dataset includes over 12.4M motion frames, significantly expanding coverage compared to earlier datasets like MoYo~\cite{tripathi20233d} and PSU-TMM100~\cite{kissos2020beyond}. 
They integrates plantar pressure with RGB-based motion capture, but still treats pressure as an auxiliary signal, underexploring its rich physical semantics. This limits its potential for contact-aware modeling and fails to leverage pressure sensing as a privacy-friendly alternative to vision-based systems.

\subsection{Controllable Motion Synthesis}

Beyond generating human motion from natural language descriptions~\cite{guo2020action2motion,petrovich2021action,t2m,petrovich2022temos,jiang2023motiongpt, motiondiffuse,tevet2023human,Mofusion,chen2024text, momask}, recent research has increasingly explored controllable motion synthesis, where additional user-defined signals such as spatial constraints are introduced to steer or constrain the synthesis process.
There are some approaches~\cite{posenetedit} that support partial pose constraints, while a prominent line of work focuses on controlling the position of key joints.
PriorMDM~\cite{shafir2024human} finetunes MDM~\cite{tevet2023human} to allow control over end-effector trajectories, while GMD~\cite{karunratanakul2023guided} and Trace and Pace~\cite{rempe2023trace} guide motion through root joint trajectories. OmniControl~\cite{omnicontrol}, MotionLCM~\cite{dai2024motionlcm}, MaskControl~\cite{Pinyoanuntapong2025MaskControl} and ACMDM~\cite{meng2025absolute} generalizes this idea, all leveraging ControlNet~\cite{zhang2023adding} for conditioning. Sketch2Anim~\cite{Zhong2025Sketch2Anim} further extends this paradigm by leveraging both ControlNet and Adapter~\cite{rebuffi2017learning} to generate motion jointly conditioning on action semantics, keyposes and trajectories.
~\cite{Hwang_2025_ICCV} takes input as sparse and flexible keyjoint signals to synthesize full-body motion.
While such methods provide flexible and expressive control, they rely on explicit, user-specified signals which are kinematic abstractions lacking physical grounding and cannot capture the continuous, frame-wise nature of real-world contact and balance phenomena.

In contrast, we propose pressure sequences as a dense, physically grounded control signal for motion synthesis. Unlike trajectories or keypoints, pressure maps directly measure the real-world interaction between the body and the ground, capturing contact, force distribution, and center-of-pressure dynamics at each frame, ensuring the generation of realistic and physically plausible motions.

\begin{figure*}[t]
\centering
\includegraphics[width=1.0\textwidth]{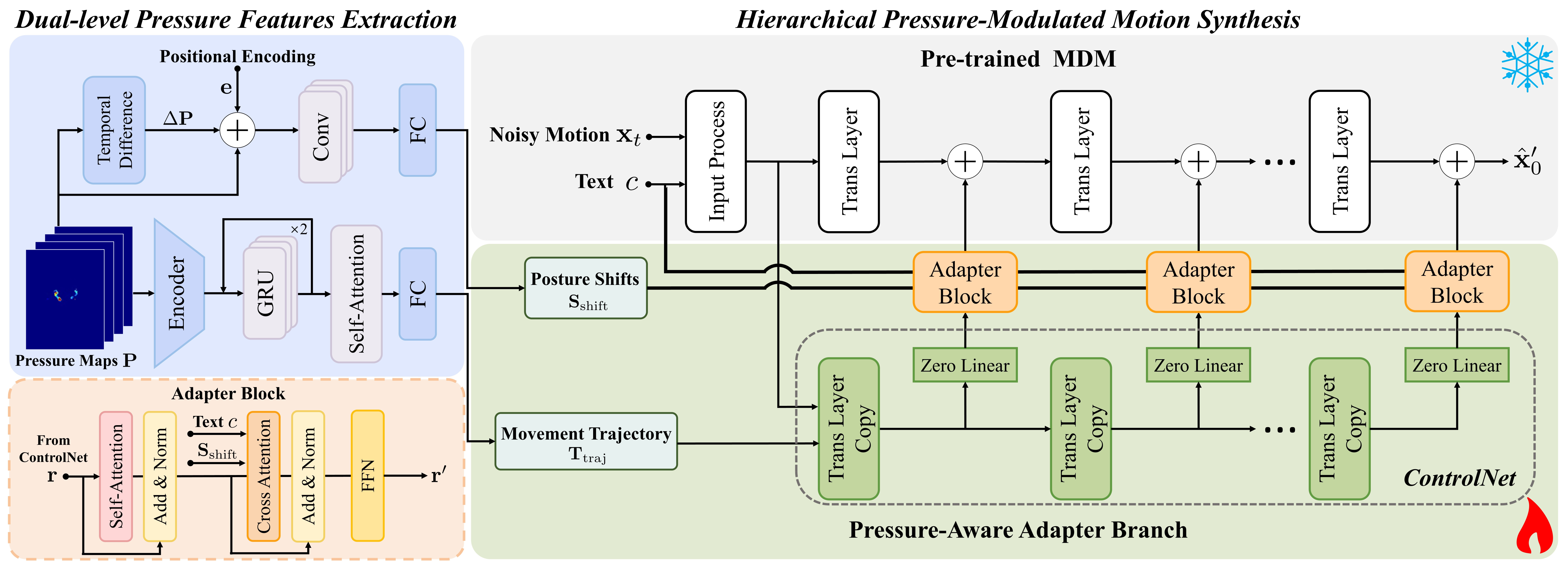} 
\caption{
The Pressure2Motion pipeline. We first extract an overall Movement Trajectory $\mathbf{T}_{\text{traj}}$ and fine-grained Posture Shifts $\mathbf{S}_{\text{shift}}$ from pressure maps. These signals are then processed by our adapter branch to provide hierarchical control: a ControlNet encodes the trajectory for global guidance, while an Adapter Block fuses this with posture shifts for local refinement. The resulting features are injected into the pretrained MDM to synthesize plausible motion aligned with the pressure signals.
}
\label{fig:pipeline}
\end{figure*}
\section{Dataset}

To explore the role of pressure signals in motion synthesis, we introduce the \dataname dataset, including Motion-Pressure-Language multimodal data for pressure-aware motion synthesis.  This dataset is constructed by expanding upon the MotionPro dataset, a large-scale motion capture dataset that records ground pressure signals alongside full-body human motion.
\dataname dataset comprises motion sequences, ground pressure data, RGB video sequences, and corresponding textual descriptions, providing a rich resource for investigating the correspondence between ground pressure signals and human motion. \cref{fig:dataset} shows samples from our dataset, including various motion sequences, their corresponding ground pressure data, and textual descriptions. 
More details can be found in the Sup. Mat.

\subsection{Dataset Statistics}
Our \dataname dataset consists of motion recordings from 25 subjects with varying heights, weights, and body types. The raw recordings are manually segmented into action-level sequences based on clear semantic boundaries, and then temporally resampled to 20 FPS. This results in a total of 20,944 motion sequences, amounting to approximately 2.3 million frames. The dataset encompasses 400 distinct categories of human motion, spanning everyday activities, traditional fitness routines, aerobic exercises, flexibility movements, and specialized motions for humanoid robotics. Each motion sequence lasts $2\sim8$ seconds, reflecting diverse temporal dynamics across action types.

In the annotation phase, each RGB image sequence is labeled using Qwen2.5-VL~\cite{Qwen2.5-VL}, a powerful vision-language model capable of processing and understanding long video sequences. For each motion sequence, we generate five textual descriptions with progressively decreasing levels of detail, resulting in a total of 104,720 descriptions across the dataset. 
By incorporating multiple descriptive levels, we enrich the variety of textual inputs, thereby enhancing our model's generalization capability and enabling it to comprehend and synthesize a diverse array of motion styles and variations.

\subsection{Motion Representation}

Unlike MotionPro, which leverages the SMPL~\cite{SMPL} parametric human model, our approach employs a more comprehensive motion representation sourced from HumanML3D~\cite{t2m}, an option that is more aptly tailored for motion synthesis frameworks.
This representation includes pelvis velocity, local joint positions, joint velocities, joint rotations (in pelvis space), and binary foot contact indicators. This format ensures fine-grained control over the motion while maintaining compatibility with existing generative models.

Specifically, during data processing, we intentionally exclude any transformations that could alter the global positions of motion joints. These transformations may lead to misalignments between the joints and the pressure data. To ensure proper alignment, we avoid such operations, keeping both the motion representation and pressure signals consistently aligned in global space, thus maintaining the integrity of the pressure-aware motion synthesis process.

\section{Method}
\label{sec:method}

Our objective is to reconstruct human motion that aligns semantically with a given textual description while ensuring physical consistency with the contact dynamics encoded in the pressure distributions. Given a sequence of pressure maps $\mathbf{P} = \{\mathbf{p}^i\}_{i=1}^{N}$ over N frames, where each $\mathbf{p}^i \in \mathbb{R}^{H \times W}$ represents a frame-wise pressure map captured by ground pressure sensors, and a corresponding text description $c$, our goal is to synthesize a temporally coherent and physically realistic motion sequence $\mathbf{x}^{1:N} = \{\mathbf{x}^i\}_{i=1}^{N}$, where $\mathbf{x}^i\in\mathbb{R}^{D}$ represents a single pose and $D$ is the dimension of the representation, reflecting both semantic intent and pressure-based physical dynamics.

In this section, we first outline the basics of the Motion Diffusion Model (MDM), then detail two core components of \methodname: Dual-level Pressure Semantics Extraction and Hierarchical Pressure-Modulated Motion Synthesis. The complete pipeline is shown in \cref{fig:pipeline}.

\subsection{Preliminary: Motion Diffusion Model}
\label{sec:preliminary}
As established in \cref{sec:intro}, reconstructing full-body motion from sparse pressure is a severely ill-posed problem. Simple regression models (as shown in \cref{tab:compare_with_baseline}) often fail, producing unrealistic or physically implausible motions. We posit that this ill-posed reconstruction task is best solved using a generative prior. Diffusion models, such as the Motion Diffusion Model (MDM)~\cite{tevet2023human}, have proven to be exceptionally powerful priors for learning complex data distributions. Therefore, we adapt MDM not as a creative tool for synthesis, but as a powerful mechanism for reconstruction, using it to find the most physically plausible motion $\hat{\mathbf{x}}_{0}$ that matches the given pressure and text prompts.

Our model builds upon and expands the capabilities of the MDM, which was initially pretrained for synthesizing motions from textual inputs, enabling its adaptation for pressure-sensitive motion synthesis.
MDM is a denoising diffusion model that follows the DDPM framework~\cite{ho2020denoising}. Given a motion sequence $\mathbf{x}$, it assumes $T$ noising steps modeled by the stochastic process:
\begin{equation}
{
q(\mathbf{x}_{t} | \mathbf{x}_{t-1}) = \mathcal{N}(\sqrt{\alpha_{t}}\mathbf{x}_{t-1},(1-\alpha_{t})I),
}
\end{equation}
for a noising step $t \in T$, where $\mathbf{x}_{T} \sim \mathcal{N}(0,I)$. MDM models the denoising process by predicting the clean motion $\hat{\mathbf{x}}_{0} = MDM(\mathbf{x}_t, t, c)$, given a noised motion $x_t$, a noise step $t$, and a textual condition $c$.
The model is trained using the standard objective:

\begin{equation}
{
\mathcal{L}_\text{diff} = E_{\mathbf{x}_0 \sim q(\mathbf{x}_0|c), t \sim [1,T]}[\| \mathbf{x}_0 - \hat{\mathbf{x}}_{0}\|_2^2].
}
\end{equation}

% New motion sampling from MDM is performed iteratively, following the DDPM framework. At each time step $t$, the clean sample $\hat{\mathbf{x}}_{0}$ is predicted and then noised back to $\mathbf{x}_{t-1}$. This process is repeated from $t=T$ until the final clean motion, $\mathbf{x}_0$, is achieved.
In our adaptation for pressure-aware motion synthesis, the pressure maps are integrated into the denoising process, providing crucial physical context that guides the synthesis of temporally consistent and physically realistic motions.

\subsection{Dual-level Pressure Features Extraction}

Extracting meaningful control signals from ground pressure images is essential for guiding the synthesized motion to align with real-world ground contact patterns and physical dynamics.
To this end, we adopt a dual-level pressure features extraction strategy that separately models Pressure-Inferred Movement Trajectory and Pressure-Induced Posture Shifts.

The Pressure-Inferred Trajectory is essential for capturing the overall movement path and body alignment during motion. To extract this information, we utilize a feature extraction module $\mathcal{F}_{\text{traj}}$ following the architecture from MotionPro~\cite{ren2025motionpro}.
This module, detailed in the Sup. Mat., uses a ResNet~\cite{he2016deep} and GRU~\cite{chung2014empirical} to process the pressure sequence.
The extracted spatial-temporal features are passed through a fully connected layer to generate a compact embedding $\mathbf{T}_{\text{traj}}$, which encapsulates the overall motion trajectory.
The extraction of the Pressure-Inferred Trajectory is represented as the encoding of pressure information from each frame:
\begin{equation}
\mathbf{T}_{\text{traj}} = \mathcal{F}_{\text{traj}}(\mathbf{P}),  \quad \mathbf{T}_{\text{traj}} = \ \{T^i\}_{i=1}^{N}.
\end{equation}
To ensure the model generalizes across various physical conditions, this module is trained separately on a diverse set of pressure-motion pairs augmented with different transformations. Once trained, the module is frozen during the motion synthesizer training phase.

In addition to capturing the overall movement trajectory, pressure maps encode detailed cues for Pressure-Induced Posture Shifts, such as center-of-mass movement, balance transitions, and subtle adjustments in posture.
To extract these dynamics, we compute both the raw pressure maps and their temporal differences, which help to capture subtle changes in posture over time, essential for synthesizing realistic and dynamic body movements. These are then combined with a grid-based positional encoding to capture spatial relationships across frames.
This combined input is passed through a module $\mathcal{F}_{\text{shift}}$, which consists of a multi-scale convolutional layer followed by a fully connected projection, resulting in a compact representation $\mathbf{S}_{\text{shift}}$ of Pressure-Induced Posture Shifts:
\begin{equation}
\mathbf{S}_{\text{shift}} = \mathcal{F}_{\text{shift}}(\mathbf{P}, \Delta \mathbf{P}, \mathbf{e}), \quad \mathbf{S}_{\text{shift}} = \{S^i\}_{i=1}^{N}, 
\end{equation}
where $\Delta \mathbf{P}$ represents the temporal difference in pressure maps and $\mathbf{e}$ is the grid-based positional encoding.
This $\mathcal{F}_{\text{shift}}$ module is trained jointly, end-to-end, with the hierarchical pressure-modulated motion synthesizer described next.

The Pressure-Inferred Movement Trajectory offers a high-level guidance for the overall path and body alignment, while the Pressure-Induced Posture Shifts capture the fine-grained, dynamic changes in posture that occur throughout the motion. Together, these features serve as the foundational control inputs, ensuring that the synthesized motion not only reflects the semantic intent expressed in the textual description but also adheres to the physical constraints dictated by the pressure signals.

\subsection{Hierarchical Pressure-Modulated Motion Synthesis}

To leverage both broad-scale trajectories and subtle posture shifts, we introduce an architecture consisting of the pre-trained MDM and the Pressure-Aware Adapter Branch, which serves as the control branch for the MDM.

The pre-trained MDM, $\mathcal{F}_\theta$, accepts as input the noisy motion $\mathbf{x}_t$, the text description $c$, and the time step $t$. It outputs the predicted clean motion, $\hat{\mathbf{x}}_0$, reflecting the semantic alignment with the given text prompt. To enhance the pressure-aware motion synthesis, the Pressure-Aware Adapter Branch is introduced. This branch consists of a ControlNet and multiple Adapter blocks that work in parallel. 

The ControlNet module $\mathcal{F}_{\text{Ctrl}}$ is implemented as a trainable variant of the pre-trained MDM, initialized with parameters from the original backbone, with a series of zero-initialized linear layers $\mathcal{Z}$ added to each layer. 
We then inject the Pressure-Inferred Trajectory embeddings $\mathbf{T}_{\text{traj}}$ directly into the noisy motion sequence $\mathbf{x}_t$ through element-wise addition. This modified input is passed through the ControlNet to produce a set of residual features $\mathbf{r}$, which are then added to the corresponding adapter blocks, guiding the synthesis process towards motion that aligns with the Pressure-Inferred Trajectory. The whole process can be written as:
\begin{equation}
\mathbf{x}_{t}' = \mathbf{x}_t + \mathbf{T}_{\text{traj}}, \quad
\mathbf{r} = \mathcal{F}_{\text{Ctrl}}(\mathbf{x}_{t}', t, c).    
\end{equation}

To address the need for leveraging Pressure-Induced Posture Shifts, which reflect subtle and localized dynamics, the Adapter blocks
$\mathcal{F}_{\text{Adapt}}$ 
are introduced. 
The ControlNet provides high-level guidance by aligning the motion with the overall movement path, while the Adapter blocks refine this motion at a local level, incorporating subtle, pressure-induced posture shifts for more detailed motion synthesis.
Operating in parallel with the ControlNet $\mathcal{F}_{\text{Ctrl}}$, 
the Adapter blocks receive residual features $\mathbf{r}$ from the $\mathcal{F}_{\text{Ctrl}}$ as input, while also integrating Pressure-Induced Posture Shift features $\mathbf{S}_{\text{shift}}$ and text embeddings $c$.
This design ensures that the high-level motion trajectory provided by the ControlNet is supplemented with fine-grained adjustments from the Adapter blocks, which captures subtle posture shifts and enhances the overall realism of the synthesized motion.
Each Adapter block consists of a self-attention layer, a cross-attention module, and a feed-forward network, producing updated residuals that are injected into the motion denoising network.

This framework enables the synthesis of motion that respects both the dynamics in the pressure signals and the semantic intent encoded in the text descriptions, ensuring hierarchical pressure-aware and semantically consistent motion synthesis.
We can get the final predicted clean motion:
\begin{equation}
\hat{\mathbf{x}}_{0}' = \mathcal{F}_\theta (\mathbf{x}_t, t, c) + \mathbf{r}' , \quad
\mathbf{r}' = \mathcal{F}_{\text{Adapt}}(\mathcal{Z}(\mathbf{r}), \mathbf{S}_{\text{shift}}, c).    
\end{equation}

Finally, to ensure a strong coupling between the generated motion and the input pressure signal, we introduce a \textbf{pressure-motion consistency loss} alongside the standard diffusion loss. 
This loss measures the alignment between 
the reconstructed motion's key joints $R(\hat{\mathbf{x}}_{0}')$% the synthesized motion and the pressure features. 
and the pressure-inferred trajectories $\mathbf{T}_{\text{traj}}$.
Since pressure primarily reflects foot contact and overall body trajectory, we compute the consistency loss based on the global positions of five key joints: the pelvis root, left/right ankle, and left/right foot. The consistency loss is given by:
\begin{equation}
\mathcal{L}_{\text{cons}}(\mathbf{T}_{\text{traj}} , \hat{\mathbf{x}}_{0}') = \frac{\sum_n \sum_j \sigma_{nj} \odot \left\lVert \mathcal{E}(\mathbf{T}_{\text{traj}}) - R(\hat{\mathbf{x}}_{0}') \right\rVert}{\sum_n \sum_j \sigma_{nj}},
\end{equation}
where $\sigma_{nj}$ is a binary value indicating whether the control signals contains a value at frame $ n $ for joint $j $. The equation for $\sigma_{nj}$ specifies that $\sigma_{nj} = 1$ when $j$ is one of the five key joints: pelvis, left ankle, right ankle, left foot, or right foot, and $\sigma_{nj} = 0$ otherwise.
$\mathcal{E}(\cdot)$ extracts control joints positions and \( R(\cdot) \) transforms the motion representation to global absolute joints positions. 
This consistency loss encourages the generated motion to be aligned with the contact patterns encoded in the input pressure maps.
Thus, the total training loss is given by the sum of the diffusion loss and the consistency loss, weighted by factors $\lambda_{\text{diff}}$ and $\lambda_{\text{cons}}$ :
\begin{equation}
\mathcal{L}_{\text{total}} = \lambda_{\text{diff}} \mathcal{L}_{\text{diff}} + \lambda_{\text{cons}} \mathcal{L}_{\text{cons}}.
\end{equation}

\section{Experiments}
\label{sec:exp}
\subsection{Experimental Settings}

\begin{figure*}[ht]
\centering
\includegraphics[width=1.0\textwidth]{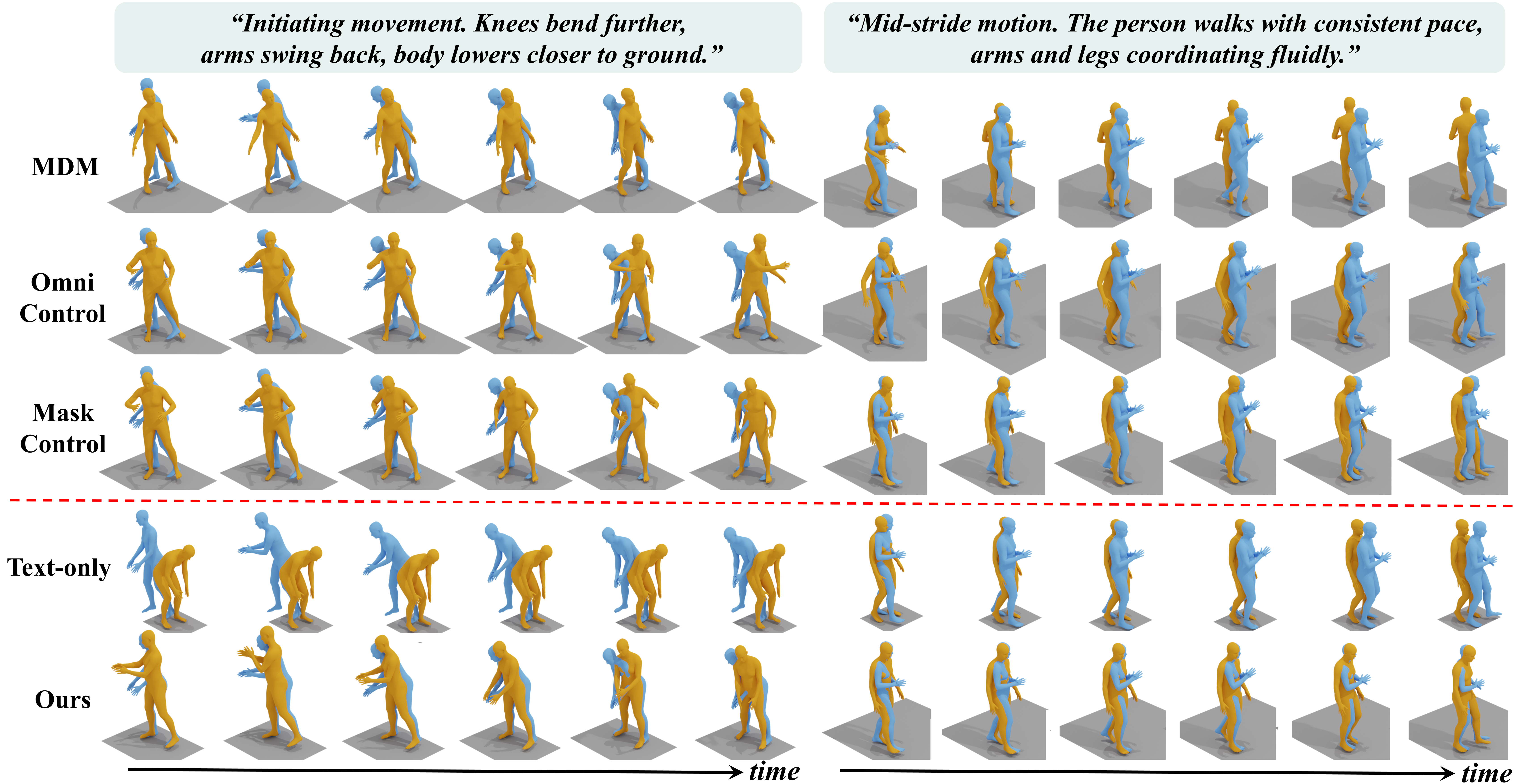} 
\caption{Visual comparisons on the MPL dataset. \expyellow{Yellow} denotes the predicted results of different methods; \expblue{blue} represents the ground-truth motions. The motions reconstructed by ours align best with the ground truth, especially in the foot region.} 
\label{fig:comparison}
\end{figure*}

\begin{table*}[ht]
\caption{Comparison of motion reconstruction with pressure control signal on the MPL dataset.}
\centering
\small

\begin{tabular}{l|cc|cccc|c}
\toprule
Method & FID $\downarrow$ & Foot Skating $\downarrow$ & CoP Error $\downarrow$ & LMPJPE $\downarrow$ & MPJPE $\downarrow$ & \begin{tabular}[c]{@{}c@{}}Trajectory Error\\ ($>$ 50cm) $\downarrow$\end{tabular} & \begin{tabular}[c]{@{}c@{}}R-precision\\ Top-3 $\uparrow$\end{tabular} \\
\midrule 
Ground Truth & 0.002 & 0.0000 & 0.0000 & 0.0000 & 0.0000 & 0.000 & 0.64 \\
\midrule
MDM\cite{tevet2023human} & 4.819 & 0.1029 & 0.9238 & 0.2550 & 0.2996 & 0.2744 & 0.458 \\
MotionDiffuse\cite{motiondiffuse} & 3.812 & 0.1138 & 0.8765 & 0.2305 & 0.2884 & 0.2650 & 0.486 \\
OmniControl\cite{omnicontrol} & 0.315 & 0.0629 & 0.5862 & 0.1362 & 0.1719 & 0.1035 & 0.523 \\
MaskControl\cite{Pinyoanuntapong2025MaskControl} & 0.388 & 0.0617 & 0.5644 & 0.1335 & 0.1695 & \textbf{0.1009} & 0.534 \\
Text-Only & 0.872 & 0.1560 & 1.0810 & 0.2320 & 0.2838 & 0.3082 & 0.2866 \\
\midrule
Regression & 40.015 & 0.7338 & 1.4832 & 0.4322 & 0.4896 & 0.5864 & 0.127 \\
\midrule
Ours & \textbf{0.262} & \textbf{0.0553} & \textbf{0.4260} & \textbf{0.1273} & \textbf{0.1622} & 0.1445 & \textbf{0.545} \\
\bottomrule
\end{tabular}

\label{tab:compare_with_baseline}
\end{table*}

\paragraph{Dataset.}
We train and evaluate our approach on the proposed \dataname dataset. Data augmentations, including random spatial translations and rotations, are implemented to enhance generalization.
The dataset is split into training, validation, and test sets with proportions of 80\%, 15\%, and 5\%. More training details can be found in the Sup. Mat.

\paragraph{Evaluation Metrics.}
We follow the evaluation protocol from OmniControl~\cite{omnicontrol}, combining motion quality from HumanML3D~\cite{t2m} and trajectory accuracy from GMD~\cite{karunratanakul2023guided}. 
As our task is a MoCap problem,
we introduce two alignment metrics—Mean Per Joint Position Error (MPJPE) and Lower-body MPJPE (LMPJPE)—to assess the reconstruction accuracy
, with LMPJPE focusing on joints relevant to foot-ground interactions. 
Additionally, to directly validate the pressure-motion consistency—a core challenge not addressed by the metrics above—we introduce the Center of Pressure Error (CoP Error). This metric measures the mean L2 distance between the Center of Pressure calculated from the input pressure map and the Center of Pressure inferred from the reconstructed lower-body joint positions. A lower CoP Error indicates superior physical alignment with the input pressure signal.
All evaluations are conducted using a motion evaluator trained on our \dataname dataset, following HumanML3D settings.

\subsection{Comparison Experiments}
As stated in \cref{sec:preliminary}, this ill-posed pressure-to-motion reconstruction task is best solved using a generative prior.  
To ensure fair comparisons and enable per-frame pressure-based control, we adapt each baseline accordingly.
We emphasize that our task is fundamentally different from controllable generation~\cite{omnicontrol, Zhong2025Sketch2Anim}. Those methods condition on kinematic abstractions (e.g., keypoints or trajectories) with a direct geometric mapping to the output. In contrast, our method must interpret a physical signal (pressure) that is sparse, noisy, and has no direct kinematic correspondence to the full-body pose, presenting a unique ill-posed reconstruction challenge.
The specific adaptation details for each baseline are provided in the Sup. Mat.

We compare our method against several baselines~\cite{tevet2023human,motiondiffuse, omnicontrol, Pinyoanuntapong2025MaskControl}, a crucial \textbf{Text-Only} baseline (our model with the pressure branch masked), and a simple \textbf{Regression} model (our diffusion model reduced to a single step). All baselines were adapted for per-frame pressure control; details are in the Sup. Mat.

The results in \cref{tab:compare_with_baseline} strongly validate our approach. Our full model achieves state-of-the-art results across reconstruction accuracy (MPJPE/LMPJPE) and realism (FID/Foot Skating). 
Crucially, our method demonstrates a substantial improvement in the new pressure-motion consistency metric (CoP Error), proving our hierarchical model uniquely learns to align the reconstructed motion with the physical input signal. 
Furthermore, the \textbf{Text-Only} baseline confirms the necessity of pressure, as its physical realism metrics (CoP Error, Foot Skating) collapse. Our full model also achieves the highest semantic alignment (R-precision), demonstrating a superior ability to balance both physical and textual constraints.

As visualized in \cref{fig:comparison}, our method produces physically realistic motions with accurate foot-ground contact, avoiding the foot-sliding and joint-misalignment artifacts common in baselines.
An analysis of intricate cases, such as motions involving no foot-ground contact (e.g., jumping) or exhibiting uncommon pressure patterns is provided in the Sup. Mat.

\begin{figure}[t]
\centering
\vspace{-0.1in}
\includegraphics[width=0.45\textwidth]{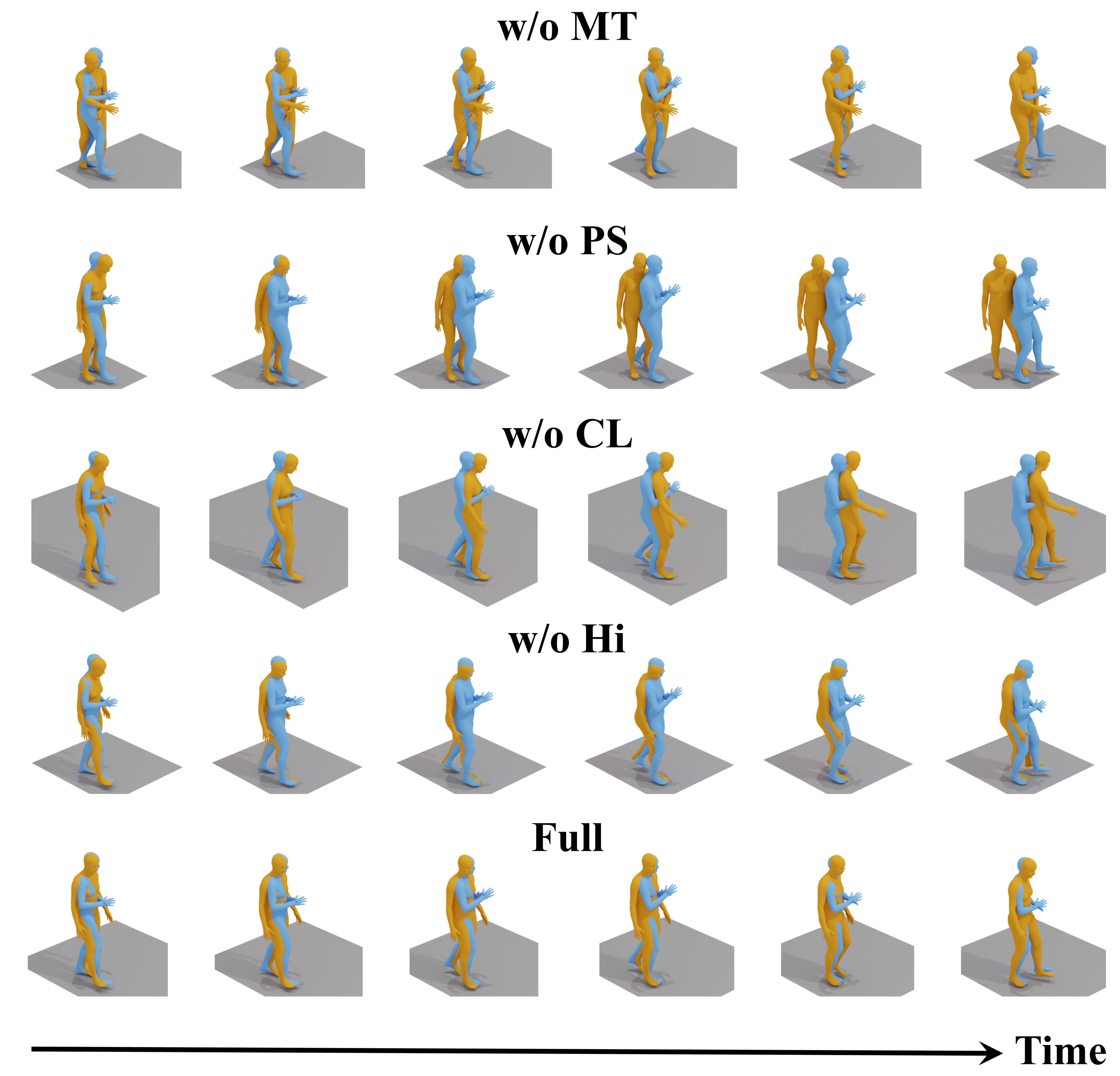} 
\caption{Visualization results of ablation study.}
\vspace{-0.1in}
\label{fig:ablation}
\end{figure}

\begin{table}[t]
\centering
\vspace{0.1in}
\caption{Ablation study of: Movement Trajectory (MT), Posture Shifts (PS), and the Consistency Loss (CL), and our Hierarchical design (Hi).}
\setlength{\tabcolsep}{1.35mm} 
\small
\begin{tabular}{l|cc|ccc}
\toprule
Method & FID$\downarrow$ & FS$\downarrow$ & CoP Err $\downarrow$ & LMPJPE$\downarrow$ & MPJPE$\downarrow$  \\
\midrule
w/o MT & 0.543 & 0.0665 & 0.8840 & 0.1943 & 0.2357 \\
w/o PS & 0.847 & 0.0629 & 0.5864 & 0.1555 & 0.2025 \\
w/o CL & 0.282 & 0.0721 & 0.5320 & 0.1550 & 0.1896 \\
w/o Hi & 0.345 & 0.0615 & 0.5610 & 0.1311 & 0.1692 \\
\midrule
Full & \textbf{0.262} & \textbf{0.0553} & \textbf{0.4260} & \textbf{0.1273} & \textbf{0.1622} \\
\bottomrule
\end{tabular}
\vspace{0.1in}
\label{tab:ablation}
\end{table}

\subsection{Ablation Study}

Our ablation study evaluates each key component, with quantitative results in \cref{tab:ablation} and qualitative examples in \cref{fig:ablation}.
As shown in \cref{tab:ablation}, removing MT, PS, or CL all lead to significant performance degradation. Finally, to validate our hierarchical design, we add a w/o Hi (Hierarchical) baseline. In this variant, we concatenate the $\mathbf{T}_{\text{traj}}$ and $\mathbf{S}_{\text{shift}}$ features into a single unified representation and feed them into the model through a single (non-hierarchical) branch. This unified model performs significantly worse than our full, hierarchical approach. This confirms that separating and hierarchically injecting the high-level trajectory (MT) and low-level posture (PS) signals is crucial for accurate reconstruction.
Our complete method outperforms all ablated versions, demonstrating that each component is crucial for reconstructing high-fidelity motion. 

\subsection{Effect of Text on Motion Reconstruction}

\begin{figure}[t]
\centering
\vspace{-0.1in}
\includegraphics[width=0.45\textwidth]{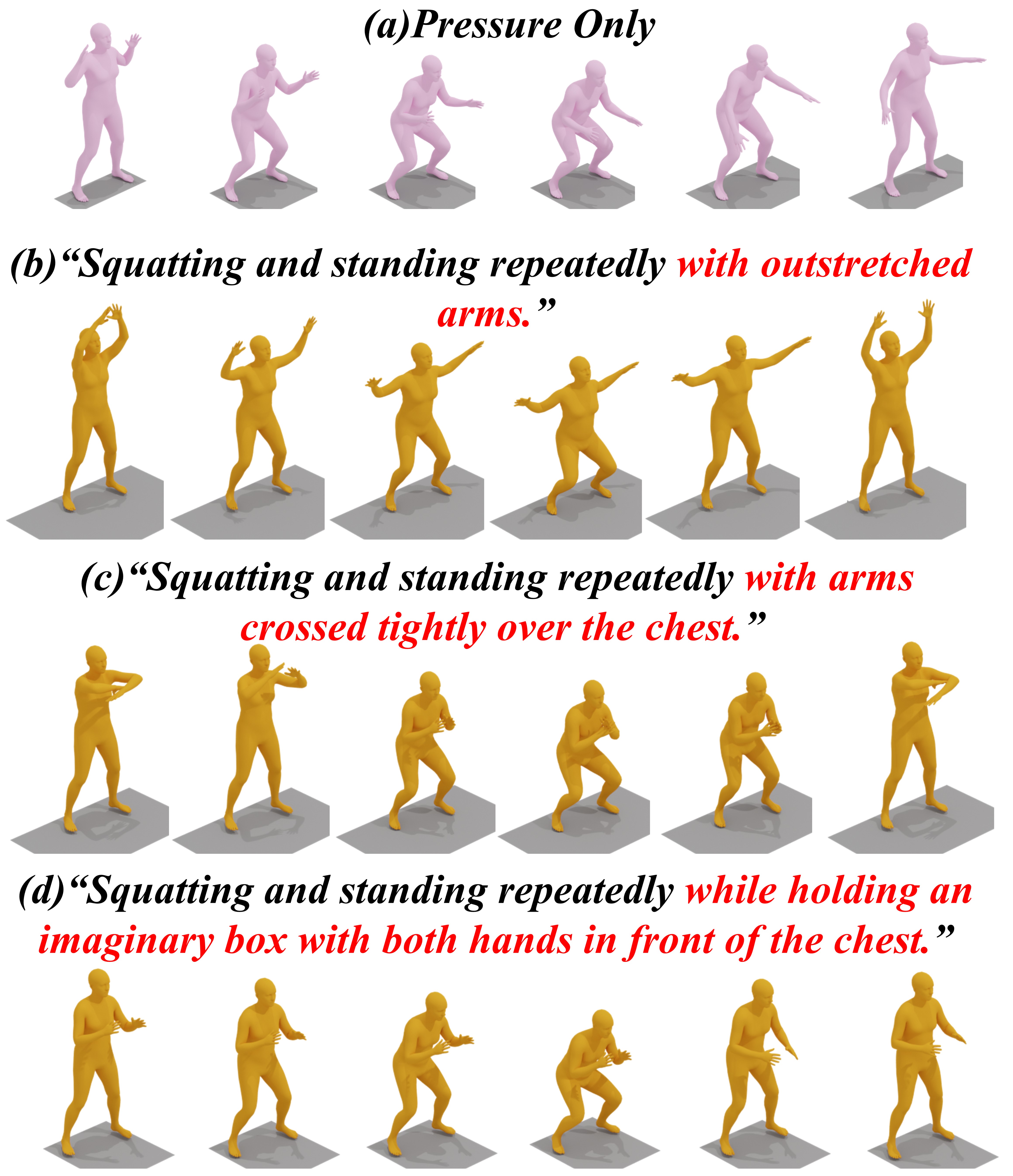}
\caption{Comparison of motion reconstruction results: (a) synthesized from pressure only, versus (b-d) synthesized from pressure with varying text prompts.}
\label{fig:text_edit}
\end{figure}

We conducted experiments to evaluate how varying textual inputs, alongside a pressure-only input condition (devoid of any textual content), influence the outcomes of motion reconstruction.
\cref{fig:text_edit} illustrates the visualization results, where (a) represents the motion synthesized with only pressure data, while (b), (c), and (d) correspond to motions synthesized with different textual descriptions. 

Specifically, the influence of text is more pronounced on the upper body, while the motion of the lower body remains closely tied to the pressure data. 
These findings validate the model's ability to reconstruct distinct motions based on varying textual descriptions while still maintaining alignment with the pressure signals.

\subsection{Real-World and OOD Generalization} 
\begin{figure}[t]
\centering
\vspace{-0.1in}
\includegraphics[width=0.45\textwidth]{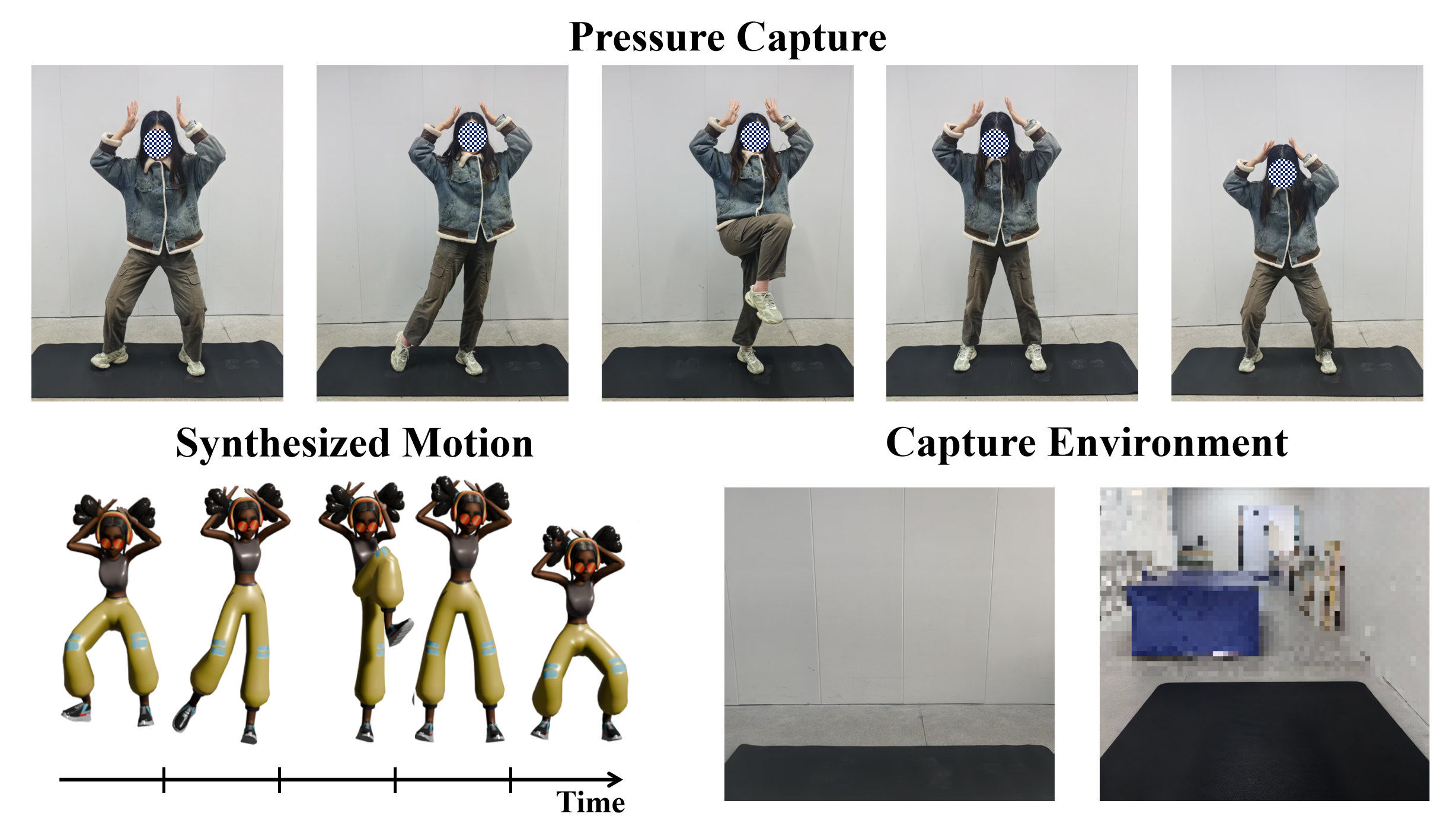} 
\caption{Real-world deployment.}
\vspace{-0.0in}
\label{fig:realworld}
\end{figure}

A critical advantage of our \modelname is its robustness and applicability in real-world, out-of-distribution (OOD) scenarios.
As illustrated in \cref{fig:realworld}, we demonstrate the model's successful generalization in an uncontrolled environment, such as a corridor.
This deployment validates that our approach maintains high-fidelity and physical realism, highlighting its potential for non-visual motion sensing in residential, clinical, or public spaces.

\section{Conclusion}
\label{sec:con}

In this work, we introduce \methodname, a pioneering approach for human motion reconstruction
from ground pressure and text prompts. Eliminating the need for cameras or wearable devices, our method enables privacy-preserving and non-intrusive motion capture. We tackle the ill-posed nature of this task with a hierarchical diffusion model specifically designed to interpret physical signals; its dual-level feature extractor decodes pressure into both broad movement trajectories and fine-grained posture adjustments. Our experiments establish strong baseline performance, demonstrating that \modelname reconstructs high-fidelity and plausible motion sequences. Furthermore, our introduction of the \dataname dataset provides the first benchmark to spur future research in this new direction.

\textbf{Limitations.}
Our primary limitations are twofold: first, the dataset is constrained to flat-surface motions, lacking complex scenarios like inclined surfaces; and second, the inference and training of out model remain computationally demanding.

{
    \small
    \bibliographystyle{ieeenat_fullname}
    \bibliography{main}
}

% WARNING: do not forget to delete the supplementary pages from your submission 
\clearpage
\setcounter{page}{1}
\maketitlesupplementary

\section{MPL Dataset Details}

\begin{figure}[!htb]
    \centering
    \begin{subfigure}[b]{0.3\textwidth}
        \centering
        \includegraphics[width=\textwidth]{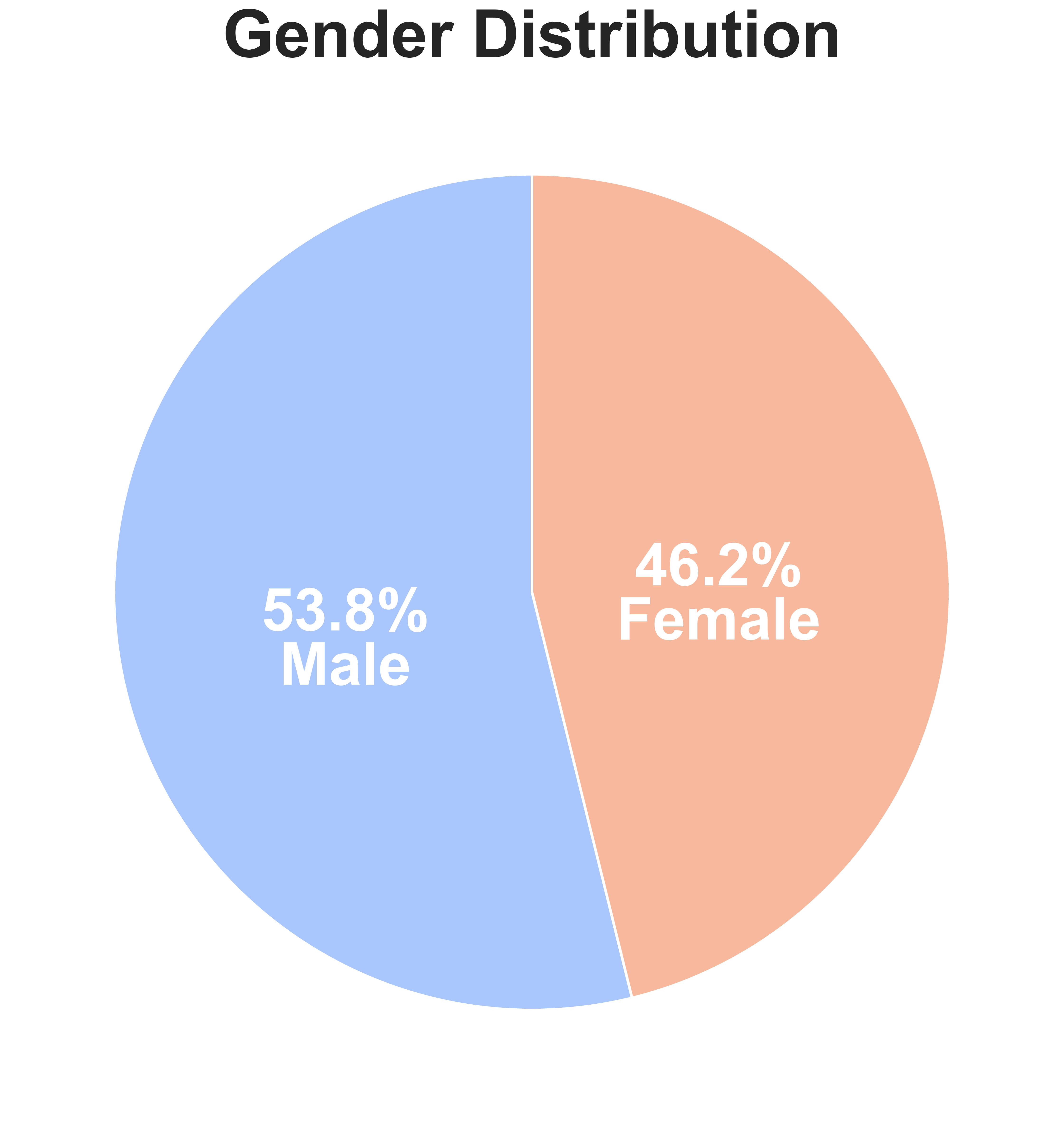}
        \caption{Gender Distribution}
        \label{fig:gender_dist}
    \end{subfigure}
    \hfill
    \begin{subfigure}[b]{0.45\textwidth}
        \centering
        \includegraphics[width=\textwidth]{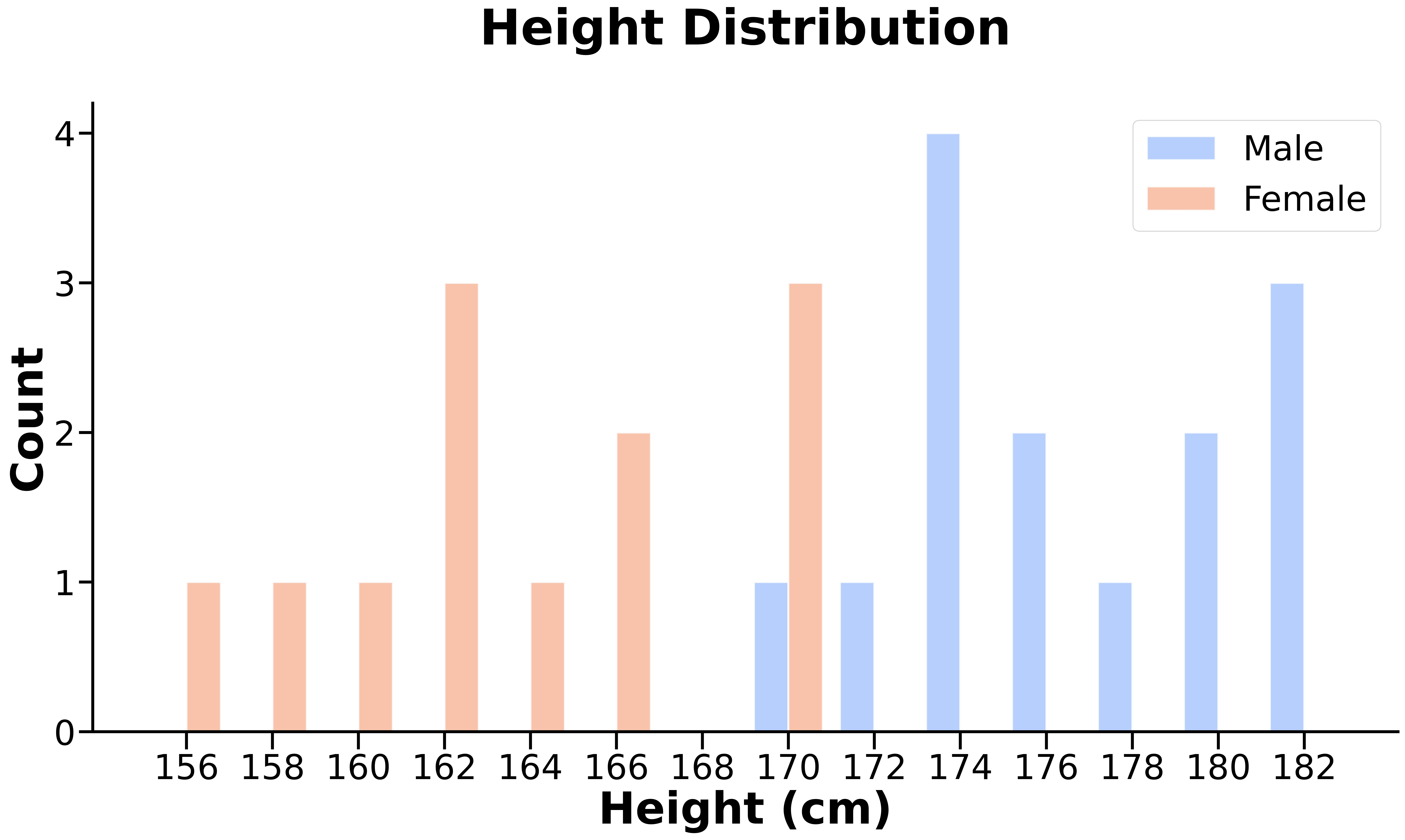}
        \caption{Height Distribution}
        \label{fig:height_dist}
    \end{subfigure}
    \hfill
    \begin{subfigure}[b]{0.45\textwidth}
        \centering
        \includegraphics[width=\textwidth]{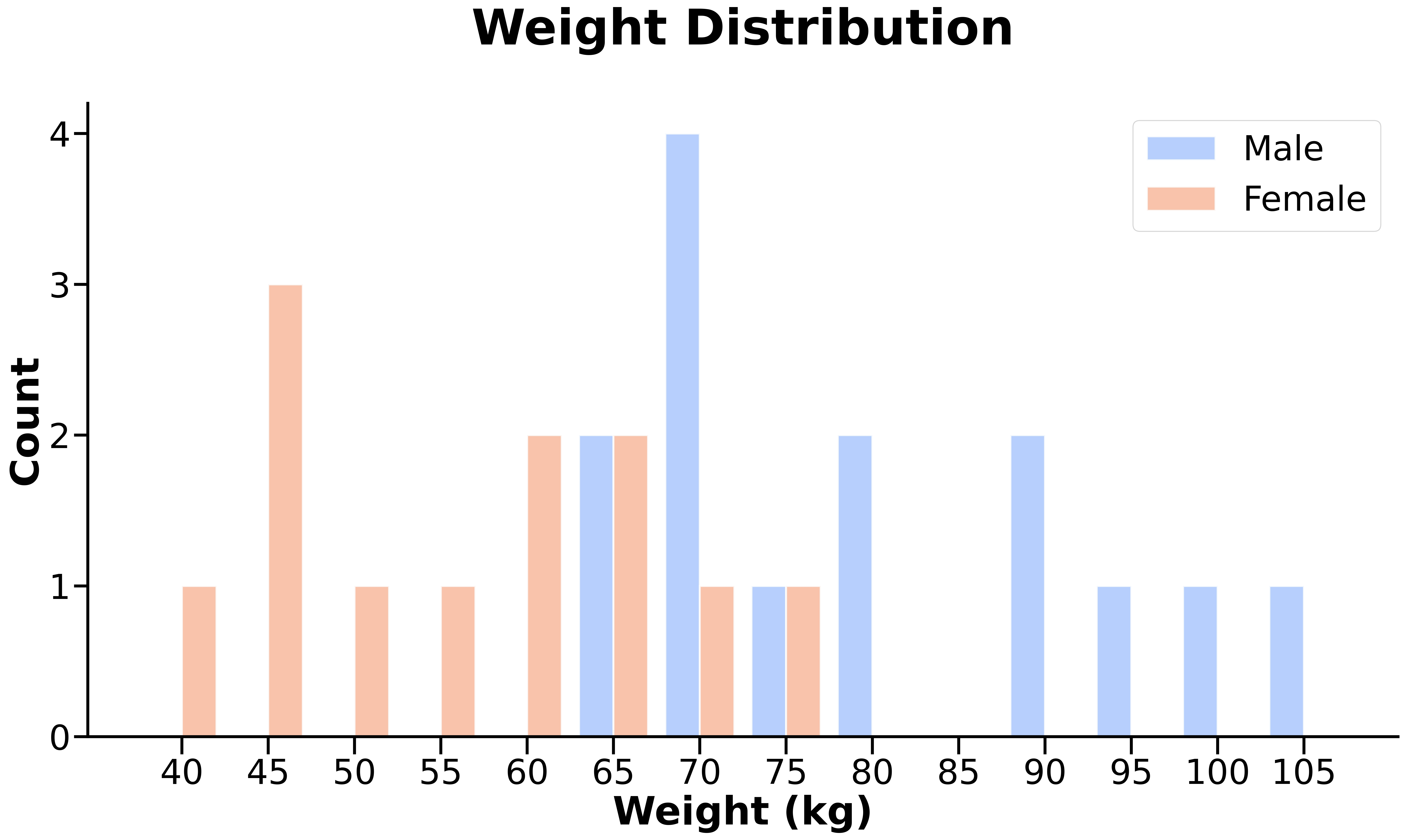}
        \caption{Weight Distribution}
        \label{fig:weight_dist}
    \end{subfigure}
    \caption{Distribution of gender, height, and weight among the 25 subjects in the \dataname dataset.}
    \label{fig:data_distribution}
\end{figure}

The \dataname dataset is developed to facilitate  research in reconstructing full-body human motion from the highly sparse inputs of ground pressure and text prompts, building on top of the MotionPRO dataset~\cite{ren2025motionpro}, which contains a large-scale collection of human motion sequences captured using plantar pressure sensors. 
For our research, we extend this dataset by incorporating textual descriptions.

The raw motion sequences in MotionPro are generally around 10 minutes long and are not segmented based on action semantics. Additionally, actions within each long sequence are often repeated 2-3 times. To address this, we manually segmented the sequences based on clear semantic action boundaries, resulting in 20,944 motion sequences, amounting to approximately 2.3 million frames. Each sequence is temporally resampled to 20 FPS for consistency, lasting $2\sim8$ seconds, reflecting diverse temporal dynamics across action types.

\subsection{Data Distribution}

Our \dataname dataset comprises motion sequences from 25 subjects with diverse physical characteristics, including a balanced distribution of gender, a wide range of heights and weights, and varying body types. 
We visualize the distribution of the following attributes:

\begin{itemize}
\item \textbf{Gender:} The dataset includes 12 females and 13 males, with a roughly balanced ratio.
\item \textbf{Height:} The subjects range from 157~$cm$ to 184~$cm$ in height, with an average of 172.1~$cm$, covering both shorter and taller individuals.
\item \textbf{Weight:} The weight distribution spans from 44.05~$kg$ to 108.00~$kg$, with an average of 56.67~$kg$, ensuring the inclusion of both lightweight and heavier subjects.
\end{itemize}

Figure~\ref{fig:data_distribution} illustrates the distribution for gender, height, and weight. This coverage enhances the robustness and applicability of our motion reconstruction model across real-world variations in body structure.

\subsection{Data Processing}

We reformatted the SMPL\cite{SMPL} parameters from the MotionPro dataset into a more comprehensive motion representation following the HumanML3D~\cite{t2m} convention. Each motion sequence of length $N$ is transformed into a representation of shape $(N, 263)$ where each frame encodes the pelvis velocity, local joint positions, joint velocities, joint rotations (in pelvis space), and binary foot contact indicators. 

During data processing, we intentionally exclude global operations such as ``uniforming skeleton" , ``put on floor,” and ``rotate to face $Z+$”, which are part of the default HumanML3D preprocessing pipeline. Uniform skeleton retargeting enforces consistent bone lengths across subjects. While these operations help standardize motion, they may distort the global positions of joints relative to the pressure data. To maintain spatial consistency, we retain the original global coordinates of both motion and pressure, ensuring accurate alignment between the two modalities during synthesis.

One notable characteristic of our processing pipeline is that the first frame of each motion sequence, i.e. the root joint of the initial pose is aligned to the origin of the $XZ$ plane. However, this causes a spatial offset between the motion and the corresponding pressure map in the $XZ$ plane. To account for this, we design a mechanism in pressure feature extraction module to predict and correct this offset using pressure information, ensuring precise spatial alignment for downstream tasks. 

\subsection{Caption Process}

Text descriptions are an integral part of the dataset, providing semantic guidance for the motion synthesis process. 
To ensure diversity and semantic richness in textual prompts, descriptions are generated using Qwen2.5-VL~\cite{Qwen2-VL}, a vision-language model capable of processing long video sequences. 
Specifically, given a motion clip and a brief action keyword from the original MotionPRO dataset, we provide the RGB video frames and keyword as input to Qwen2.5-VL. 
The model interprets the human activity within the video context and generates five diverse captions at varying levels of detail. These descriptions range from simple high-level actions (e.g., “The person is walking”) to more intricate and detailed descriptions (e.g., “The person is walking with a slight leftward tilt and right arm movement”).

\subsection{Augmentation}

Specifically, given a pressure sequence, 
we apply spatial augmentations (translations and rotations) to simulate real-world variations in global orientation and position. 
We adjust the global offset of the motion accordingly to maintain spatial alignment with the augmented pressure data.
This augmentation helps to simulate real-world variations in body posture and pressure signals.

\section{Implementation Details}

\subsection{Training Setup}
Our models are implemented in PyTorch and trained on 8 NVIDIA A800 GPUs for a total of 100,000 iterations. We adopt the AdamW optimizer~\cite{loshchilov2017decoupled} with a learning rate of $1 \times 10^{-5}$. The ControlNet is initialized with pretrained weights from MDM~\cite{tevet2023human}. During training, the parameters of the Movement Trajectory extraction module $\mathcal{F}_{\text{traj}}$ and the pretrained MDM backbone $\mathcal{F}_\theta$ are frozen to retain their original representations.

\subsection{Network and Feature Dimensions}
We follow prior works and use CLIP~\cite{radford2021learning} to encode text prompts into 512-dimensional embeddings. The output features of both the ControlNet and the Adapter modules are also of size 512 to ensure compatibility with the pretrained MDM architecture.

The Pressure-Inferred Movement Trajectory $\mathbf{T}_{\text{traj}}$ and Pressure-Induced Posture Shifts $\mathbf{S}_{\text{shift}}$ are extracted with output dimensions of $(B, L, 39)$ and $(B, L, 256)$ respectively, where $B$ is the batch size and $L = 196$ is the sequence length. The 39-dimensional trajectory representation includes the global 3D positions ($XYZ$) of the root, left/right ankles, and left/right toes (total 5 joints × 3 = 15), as well as 6D rotation representations for the left/right ankles and toes (4 joints $\times$ 6 = 24).

\subsection{Diffusion and Loss Hyperparameters}
To improve robustness to text variations, we randomly drop 10\% of the text conditions during training. This enables the use of Classifier-Free Guidance (CFG)~\cite{ho2022classifier} during inference, where we apply a CFG scale of 5.
We adopt a standard DDPM~\cite{ho2020denoising} framework with $T = 1000$ denoising steps. The control strength $\tau$ for injecting pressure signals is defined as $\tau = \frac{20 \hat{\Sigma}_t}{L}$, where $\hat{\Sigma}_t = \min(\Sigma_t, 0.01)$.
We set $\lambda_{\text{diff}} = 1$ and $\lambda_{\text{cons}} = 5$ throughout all experiments.

% baselines
\subsection{Baseline Adaptations} 

For MDM\cite{tevet2023human} and MotionDiffuse\cite{motiondiffuse}, we concatenate the global and local pressure embeddings and append them to the noisy motion input at each denoising step. This allows these models to incorporate pressure signals at each frame, providing a consistent pressure-aware motion reconstruction. For OmniControl\cite{omnicontrol} and MaskControl\cite{Pinyoanuntapong2025MaskControl}, we replace the original spatial control inputs with the combined pressure embeddings, enabling these models to condition on pressure in a comparable manner to our approach.

% model
\subsection{Pressure Feature Extractor Details}

The Pressure-Inferred Trajectory $\mathcal{F}_{\text{traj}}$ is essential for capturing the overall movement path and body alignment. To extract this information, we utilize a feature extraction module following the architecture from MotionPro~\cite{ren2025motionpro}, which includes a ResNet-based~\cite{he2016deep} pressure encoder, a temporal information processor, and a fully connected projection layer. Given the sparsity of pressure maps, where valid values are limited and primarily found under the feet during standing, the pressure encoder utilizes a compact ResNet architecture with small convolutional kernels to focus on the localized pressure regions, despite the large size of the pressure map. 
Temporal dynamics are captured using the temporal information processor, which combines a GRU~\cite{chung2014empirical} to model long-term dependencies with a self-attention mechanism to capture short-term correlations in the pressure sequence.

\section{Evaluation Details}

We adopt a text feature extractor and a motion feature extractor from HumanML3D and retrain it on our \dataname dataset to adapt to the new data distribution. The resulting model is used to evaluate all the methods.

We evaluate motion quality, motion-pressure consistency, and semantic alignment of the reconstructed
motions using the following metrics:

\begin{itemize}
    \item \textbf{Center of Pressure (CoP) Error $\downarrow$:}
    This metric directly measures the physical consistency between the input pressure signal and the reconstructed motion. It is calculated as the mean L2 distance between two CoP time-series:
    
    \textbf{Pressure CoP ($CoP_{Pressure}$):} Calculated from the input pressure map $P_n \in \mathbb{R}^{H \times W}$ at frame $n$. The pixel-space CoP (geometric center) is computed as a weighted average:
    $$
    CoP_{P, x}^{(n)} = \frac{\sum_{i,j} P_n(i, j) \cdot j}{\sum_{i,j} P_n(i, j)}, 
    $$
    $$
    CoP_{P, z}^{(n)} = \frac{\sum_{i,j} P_n(i, j) \cdot i}{\sum_{i,j} P_n(i, j)}.
    $$
    This pixel-space CoP is then transformed into motion-space using a pre-calibrated scale $S$ and offset $O$: 
    $$CoP_{Pressure}^{(n)} = [CoP_{P, x}^{(n)}, 0, CoP_{P, z}^{(n)}] \odot S + O.
    $$

    \textbf{Motion CoP ($CoP_{Motion}$):} Inferred from the reconstructed motion's lower-body joints. We use a softmax-weighted average of the $K$ key foot joints' (e.g., ankles, toes) ground projections $j_k = [j_{k,x}, j_{k,y}, j_{k,z}]$, where the weight $w_k$ is inversely related to the joint's height $j_{k,y}$:
    $$
    w_k^{(n)} = \frac{\exp(-j_{k,y}^{(n)} / \tau)}{\sum_{l=1}^{K} \exp(-j_{l,y}^{(n)} / \tau)},
    $$
    $$
    CoP_{Motion}^{(n)} = \sum_{k=1}^{K} w_k^{(n)} \cdot [j_{k,x}^{(n)}, 0, j_{k,z}^{(n)}].
    $$

    \textbf{CoP Error ($\mathcal{L}_{CoP}$):} The final error is the mean Euclidean distance over all $N$ frames and $B$ batch items:
    $$
    \mathcal{L}_{CoP} = \frac{1}{B \cdot N} \sum_{b=1}^{B} \sum_{n=1}^{N} \left\| CoP_{Pressure}^{(b,n)} - CoP_{Motion}^{(b,n)} \right\|_2.
    $$
    A lower value indicates superior motion-pressure consistency.
    
    \item \textbf{Fréchet Inception Distance (FID) $\downarrow$:}
    FID measures the distributional distance between reconstructed motions and ground-truth motions in the feature space.  
    In our setting, motions are encoded via a pre-trained motion encoder, and FID is computed on the extracted features. Lower FID indicates that the reconstructed motions are more realistic and distributionally similar to real data.

    \item \textbf{Foot Skating $\downarrow$:}  
    This metric computes the ratio of frames in which a foot joint is supposed to be in contact with the ground but exhibits non-negligible motion, indicating physically implausible sliding. Specifically, for each frame, we check whether a foot is labeled as "in contact" and simultaneously has a velocity exceeding a small threshold. The ratio of such inconsistencies over all frames is reported. A lower value indicates better foot-ground contact realism and physical plausibility.

    \item \textbf{Mean Per Joint Position Error (MPJPE) $\downarrow$:}  
    MPJPE measures the average Euclidean distance between corresponding joints in the predicted and ground-truth motions:
    \[
    \text{MPJPE} = \frac{1}{N \cdot T} \sum_{t=1}^{T} \sum_{j=1}^{N} \| \hat{\mathbf{p}}_{t,j} - \mathbf{p}_{t,j} \|_2
    \]
    where $T$ is the number of frames, $N$ is the number of joints, and $\hat{\mathbf{p}}_{t,j}$ and $\mathbf{p}_{t,j}$ denote the predicted and ground-truth positions of joint $j$ at time $t$.
    This metric evaluates the spatial alignment between reconstructed and real motions.

    \item \textbf{Lower-body MPJPE (L-MPJPE) $\downarrow$:}  
    A variant of MPJPE that only considers lower-body joints (e.g., hips, knees, ankles, feet), which are most relevant to pressure-ground interactions. It reflects the model's ability to reconstruct physically grounded lower-body motion. Lower is better.

    \item \textbf{Trajectory Error ($>$ 50cm) $\downarrow$:}  
    This metric measures the ratio of motion sequences in which trajectory frames deviate from the ground truth by more than 50 cm. It reflects whether global body movement is consistently aligned with the physical signal.

    \item \textbf{R-Precision (Top-3) $\uparrow$:}  
    R-Precision measures the semantic consistency between reconstructed motions and their associated text prompts. We use a joint motion-text encoder to compute the similarity between reconstructed motion features and ground-truth text embeddings. R-Precision@3 reflects whether the correct caption is ranked among the top-3 retrieved results for a reconstructed motion. Higher values indicate better semantic alignment.

\end{itemize}

\section{Additional Ablation Study}

In addition to the ablation study on pressure features, we further investigate the impact of model architecture by removing key components, namely the ControlNet and Adapter. Specifically, we modify the architecture by concatenating the two pressure features—Movement Trajectories and Posture Shifts—directly and feeding them into either the ControlNet or Adapter without using the hierarchical structure.

The results of this experiment are shown in Table~\ref{tab:add_ablation}, where we compare the full model with versions that exclude ControlNet and Adapter. For the version without ControlNet, we observe a significant increase in FID (1.3683) and MPJPE (0.1951), indicating that removing the ControlNet impairs the model's ability to properly align the motion with the pressure signals, resulting in less accurate motion reconstruction. Similarly, removing the Adapter leads to a noticeable degradation in performance, with FID increasing to 0.695 and MPJPE rising to 0.2092. These results demonstrate the critical role of both components in ensuring the high fidelity and physical plausibility of the reconstructed motions. Moreover, concatenating the two pressure features (Movement Trajectories and Posture Shifts) directly and feeding them into either the ControlNet or Adapter results in inferior performance compared to the full model. This suggests that the hierarchical structure, where ControlNet handles the overall movement trajectory and the Adapter fine-tunes the posture shifts, is essential for reconstructing realistic and semantically aligned motions.

\begin{figure*}[htp]
\centering
\includegraphics[width=0.7\textwidth]{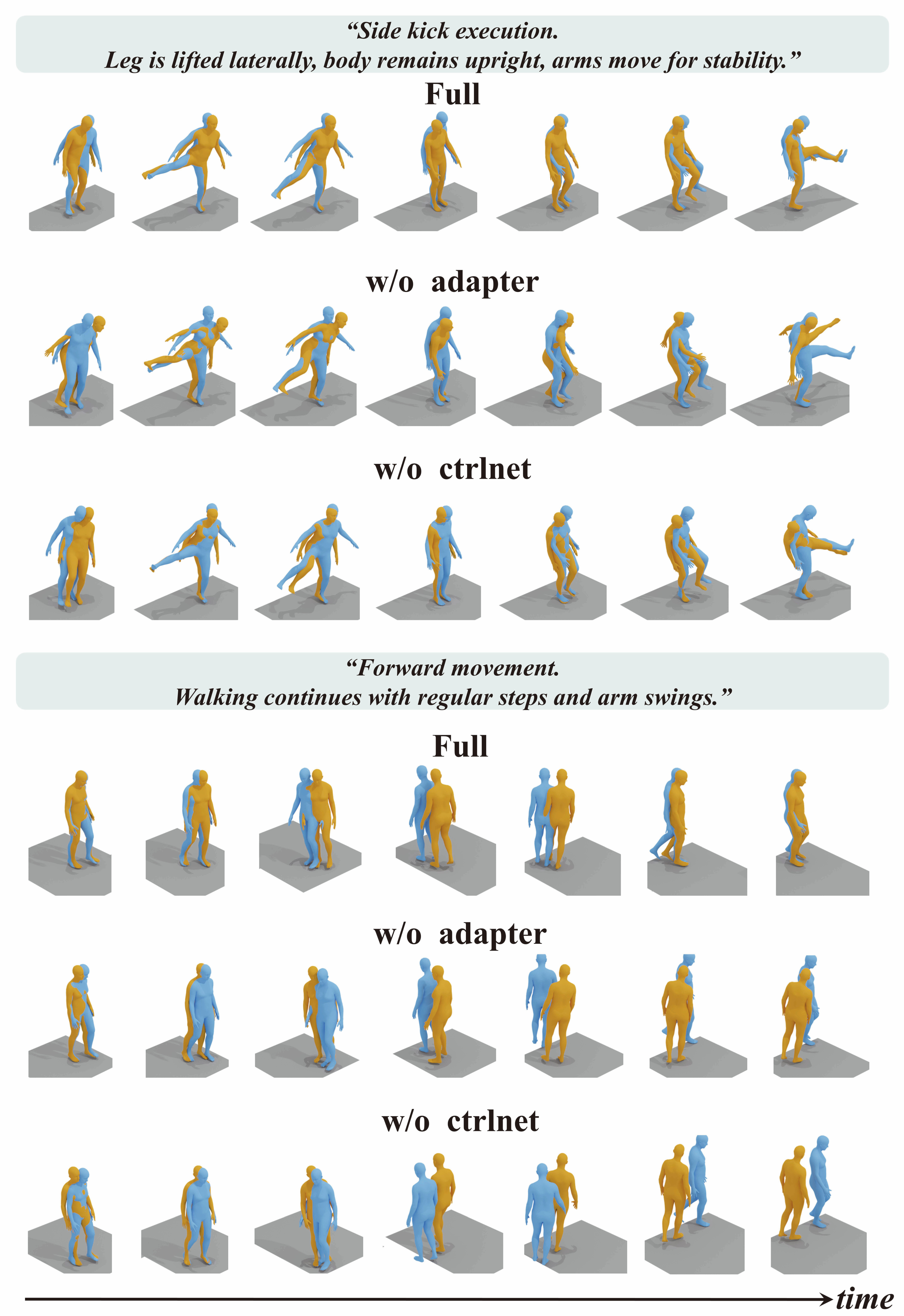} 
\caption{Additional ablation results on the MPL dataset. \expyellow{Yellow} denotes the predicted results of different methods; \expblue{blue} represents the ground-truth motions.}
\label{fig:add_ablation}
\end{figure*}

Figure~\ref{fig:add_ablation} provides visual comparisons of the motion sequences reconstructed by the different model variations. These visualizations further confirm that the full model consistently produces motions that are more physically plausible and aligned with the pressure data, especially in areas such as foot-ground contact.

\begin{table}[ht]
\caption{Additional Ablation study of Model Archtecture.}
\centering
\setlength{\tabcolsep}{1.35mm} 
\small
\begin{tabular}{l|cc|ccc}
\toprule
Method & FID$\downarrow$ & FS$\downarrow$ & Cop Err$\downarrow$ & L-MPJPE$\downarrow$ & MPJPE$\downarrow$  \\
\midrule
w/o ctrlnet & 1.3683 & 0.0621 & 0.6120 & 0.1694 & 0.1951  \\
w/o Adapter & 0.695 & 0.0634 & 0.6655 & 0.1702 & 0.2092  \\
Full & \textbf{0.262} & \textbf{0.0553} & \textbf{0.4260}& \textbf{0.1273} & \textbf{0.1622}  \\
\bottomrule
\end{tabular}

\label{tab:add_ablation}
\end{table}

\section{More Visualization Results}

\begin{figure*}[!htp]
\centering
\includegraphics[width=0.8\textwidth]{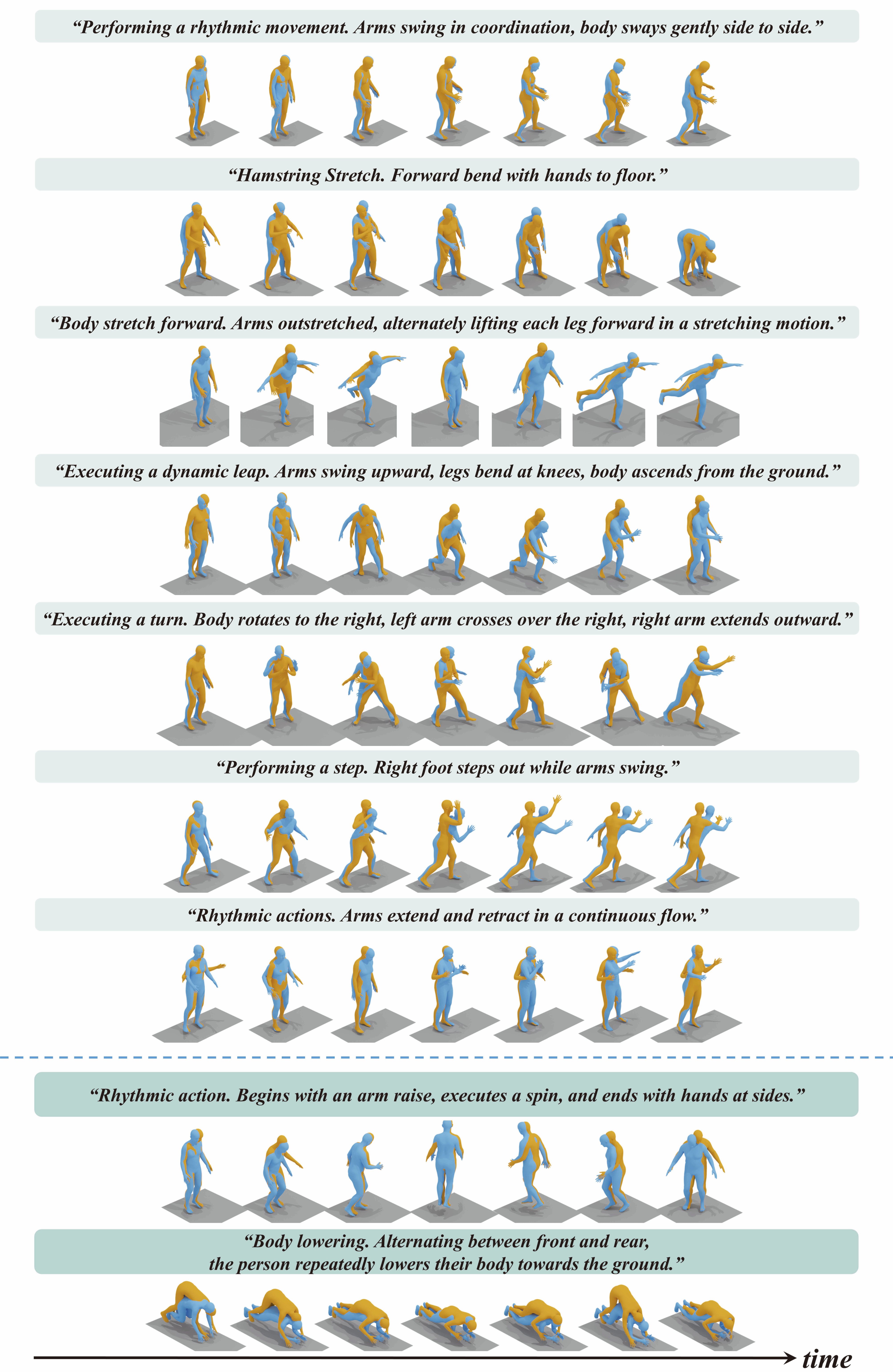} 
\caption{More visualization results on the MPL dataset. \expyellow{Yellow} denotes the predicted results of different methods; \expblue{blue} represents the ground-truth motions.} 
\label{fig:more_visual}
\end{figure*}

We present more motion reconstruction results to further demonstrate the effectiveness of our method in reconstructing human motion from sparse pressure data. Figure~\ref{fig:more_visual} showcases a variety of reconstructed motions across different scenarios, highlighting the robustness of our approach in the pressure-to-motion task.
The first few cases demonstrate typical human actions, such as walking, standing, and some daily activities. Our model reconstructs high-fidelity motions that align well with the pressure data and text prompts, maintaining both physical consistency and semantic plausibility.

Particularly interesting are the last two cases in Figure~\ref{fig:more_visual}. The second-to-last case corresponds to a jumping motion, where \textbf{no pressure} is applied during the jump. Despite the absence of pressure in the air, our model successfully reconstructs a realistic jumping motion, demonstrating its ability to handle scenarios with \textbf{no foot-ground contact.}

The final case in Figure~\ref{fig:more_visual} shows a dynamic plank position, which involves complex pressure distributions from both the hands and feet. Our method effectively handles this rare and specialized pressure contact scenario, reconstructing a physically plausible motion that corresponds to the simultaneous pressure from all four limbs. This demonstrates the versatility of our model in \textbf{handling uncommon or intricate pressure patterns.}

\section{Limitations}

Despite the promising results, our approach is subject to several limitations.

First, the diversity of motion types in the dataset remains relatively limited. While the \dataname dataset includes various basic motions such as walking, standing, and sitting, more complex activities—such as interactions on inclined surfaces or with dynamic real-world environments—are not yet covered. 
Extending the dataset to include more diverse and complex motion types, as well as scenarios involving pressure data from inclined or interacting surfaces (e.g., walking on stairs or engaging with objects), would significantly enhance the robustness and applicability of the model in real-world use cases.

Another major limitation of our model lies in its computational complexity. While our approach demonstrates high fidelity in reconstructing pressure-aware motions, the underlying architecture—specifically the pressure feature extraction module and the hierarchical pressure-modulated motion reconstruction framework—is relatively large and computationally demanding. 
Additionally, the motion reconstruction process, based on the diffusion model, involves multiple denoising steps, making the inference process slower.
On a single NVIDIA A800 GPU, reconstructing one motion sequence takes approximately 180 seconds. This extended inference time can become a bottleneck when deploying the model in real-time applications. As a future direction, we plan to investigate strategies, such as more efficient pressure feature extraction techniques, and explore inference optimization to speed up the motion reconstruction process while maintaining high-quality results.

A third limitation concerns the nature of our textual guidance. The \dataname dataset's text descriptions are entirely generated by a VLM. While this provides consistent and detailed annotations at scale, VLM-generated text tends to be homogeneous and overly descriptive, adhering to a specific stylistic pattern (e.Sg., "The person raises their left arm"). This clean, literal data distribution does not reflect the full diversity and ambiguity of real-world human language. Human prompts are often more abstract, colloquial, underspecified, or describe high-level goals rather than explicit kinematics (e.g., "Look for something on the floor" vs. "Bend over and turn head"). Consequently, our model may be overfitted to this VLM-specific text style and less robust to "in-the-wild" human-authored prompts.

Future work will focus on addressing these limitations by: 1) Enhancing robustness and expanding the dataset to include more dynamic activities (e.g., inclined surfaces); 2) Investigating inference optimization strategies to enable real-time applications; 3) Improving robustness to diverse, human-authored text; and 4) Leveraging pressure data in simulated reinforcement learning or personalized motion modeling.

\clearpage
\clearpage
% {
%     \small
%     \bibliographystyle{ieeenat_fullname}
%     \bibliography{main}
% }
\end{document}